%% file: main.tex
\documentclass[11pt,a4paper]{article}
\usepackage[affil-it]{authblk}
\usepackage{blindtext}
\usepackage{emnlp2020}
\usepackage{times}
\usepackage{soul}
\usepackage{xcolor}
\usepackage{latexsym}
\usepackage{graphicx}
\usepackage{booktabs}
\usepackage{multirow}
\usepackage{placeins}
\usepackage{tablefootnote}
\usepackage{threeparttable}
\usepackage{diagbox}
\usepackage{ulem}
\usepackage{mwe}

\newcommand{\clarify}[1]{\textcolor{red}{\textbf{Do we want to keep this?}}}

\newcommand{\newlyadded}[1]{\textcolor{red}{#1}}

\newcommand{\hide}[1]{}
\usepackage{color}
\usepackage{array}
\usepackage{etoolbox}
\usepackage{url}
\definecolor{blue}{RGB}{0, 93, 170}			
\definecolor{darkgreen}{RGB}{0, 102, 0}
\usepackage{microtype}

\hypersetup{breaklinks=true}

\newcounter{rowcntr}[table]
\renewcommand{\therowcntr}{\thetable.\arabic{rowcntr}}

\newcolumntype{N}{>{\refstepcounter{rowcntr}\therowcntr}c}

\AtBeginEnvironment{tabular}{\setcounter{rowcntr}{0}}

\aclfinalcopy
\makeatletter
\newcommand{\printfnsymbol}[1]{%
  \textsuperscript{\@fnsymbol{#1}}%
}
\makeatother
\title{Looking Beyond Sentence-Level Natural Language Inference for Downstream Tasks
}
\author[1]{Anshuman Mishra \thanks{~~Equal contribution.}~} 
\author[1]{Dhruvesh Patel\printfnsymbol{1}}
\author[1]{Aparna Vijayakumar\printfnsymbol{1}}
\author[1]{\\Xiang Li}
\author[2]{Pavan Kapanipathi}
\author[2]{Kartik Talamadupula}
\affil[1]{ College of Information and Computer Sciences,   University of Massachusetts Amherst}
\affil[2]{IBM Research}

\date{}

\begin{document}
\maketitle

\normalem 
\begin{abstract}
\input{Abstract/abstract.tex}
\end{abstract}

\section{Introduction }

\input{Intro/introduction}

\section{Related Work}
\input{"Related Work/related_work"}

\section{NLI for Downstream Tasks}
\label{sec:nli_for_downstream}

\input{"NLI Models for Downstream Tasks/nli_for_downstream_NLP"}

    \subsection{The Long Premise Conjecture}
    \label{subsec:long_premise_conjecture}
    \input{"NLI Models for Downstream Tasks/long_premise_conjecture"}

\section{Reformatting MCRC to NLI}
\label{sec:conversion}
\input{"Reformatting QA to NLI/reformatting_qa_to_nli"}

    \subsection{Rule-based Conversion}
    \label{subsec:conversion_rule}
    \input{"Reformatting QA to NLI/rule"}
    
    \subsection{Neural Conversion}
    \label{subsec:conversion_neural}
    \input{"Reformatting QA to NLI/neural"}
    
    \subsection{Hybrid Conversion}
    \label{subsec:conversion_hybrid}
    \input{"Reformatting QA to NLI/hybrid"}
    
\section{A Transferable NLI model}
    \label{sec:transferable_modeling}
    \input{"A Transferable NLI Model/transferable_modeling"}

    \section{Datasets}
    \label{sec:diverse_qa_datasets}

\input{Datasets/datasets}

\section{Experiments and Results}
\label{sec:results_and_discussion}
\input{"Results and Discussion/results_and_discussion"}

    \subsection{Evaluating Conversion Quality}
    \label{subsec:results_qa_vs_nli}
    \input{"Results and Discussion/qa_vs_nli"}

    \subsection{Long Premise Conjecture}
    \label{subsec:results_long_premise}
    \input{"Results and Discussion/LPC"}
        \subsubsection{Evaluation on MCRC}
\input{"Results and Discussion/LPC_MCRC"}

        \subsubsection{Evaluation on CFCS}
        \label{subsubsec:results_cfcs}
        \input{"Results and Discussion/LPC_CFCS"}

    

\section{Conclusion}
\label{sec:conclusion}
\input{Conclusion/conclusion}



\FloatBarrier
\bibliography{bibliography.bib}
\bibliographystyle{acl_natbib}

\FloatBarrier
\appendix

\include{appendix/appendices}

\end{document}

%% file: Abstract/abstract.tex
In recent years, the Natural Language Inference (NLI) task has garnered significant attention, with new datasets and models achieving near human-level performance on it. However, the full promise of NLI -- particularly that it learns knowledge that should be generalizable to other downstream NLP tasks -- has not been realized. In this paper, we study this unfulfilled promise from the lens of two downstream tasks: question answering (QA), and text summarization. We conjecture that a key difference between the NLI datasets and these downstream tasks concerns the length of the premise; and that creating new long premise NLI datasets out of existing QA datasets is a promising avenue for training a truly generalizable NLI model. We validate our conjecture by showing competitive results on the task of QA and obtaining the best reported results on the task of Checking Factual Correctness of Summaries.

%% file: Intro/introduction.tex

 Natural Language Inference (NLI)
 is the task of determining the relation between a given premise-hypothesis text pair; and is critical for natural language understanding.
 The availability of large-scale, open NLI datasets~\cite{Bowman2015,mnli} has recently resulted in the development of bigger and more robust models for solving the task of NLI. As some of these models close in on human-level performance, a natural question arises: {\em can models trained on these large-scale NLI datasets be used for other downstream NLP tasks?} So far, efforts towards using NLI for downstream tasks have had limited success~\cite{Trivedi2019,falke2019ranking,clark2018think}.
 
 \begin{figure}[!ht]
    \centering
    \includegraphics[width=\columnwidth]{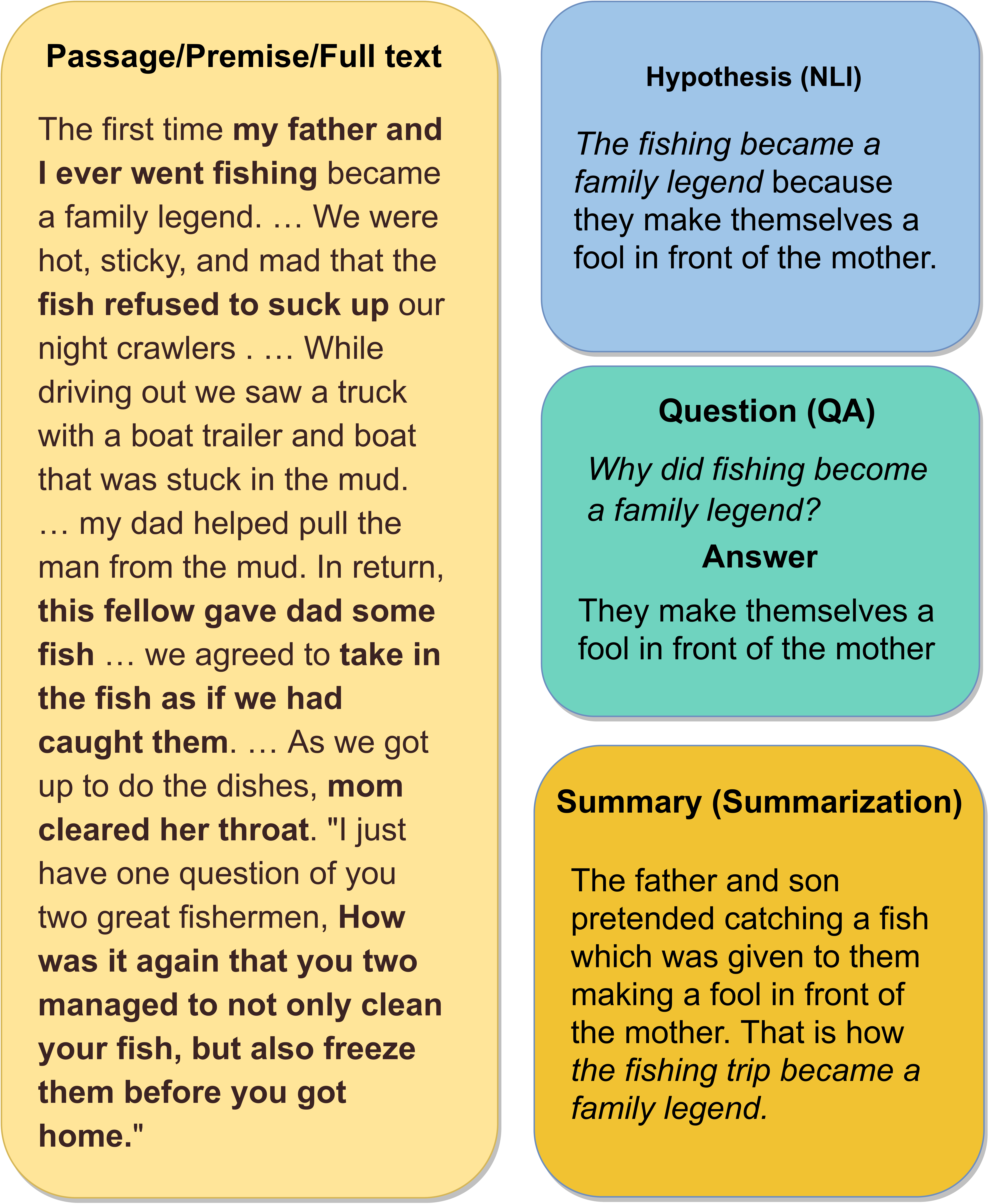}
    \caption{The tasks of Question Answering and Checking Factual Consistency of Text-Summaries can naturally be transformed into the Natural Language Inference problem.}
    \label{fig:intro_fig}
\end{figure}

One potential reason for this lack of success may be the inherent nature of the existing NLI datasets. Specifically, most existing NLI datasets consider one or at most a few sentences as the premise; and hence, can be tackled successfully by models that possess an understanding of only {\em local} sentence-specific semantics (negation, quantification, conditionals, monotonicity, etc.). On the other hand, most downstream NLP tasks of interest such as Question Answering (QA) and Text-Summarization require reasoning over much longer texts. 
While it has been posited that the capabilities required for handling sentence-level inference are very different from those required to perform inference on longer forms of text~\cite{cooper1996usingsemantic,lai-etal-2017-natural}, the effect of this on downstream tasks has not been studied.

In order to investigate this, we need to evaluate existing models -- trained on sentence-level NLI datasets -- on datasets that feature NLI instances with longer premises. However, current NLI datasets do not exhibit long premises. QA datasets~\cite{rajpurkar2016squad,lai2017race,multirc,sun2019dream,huang2019cosmos} on the other hand, encompass a variety of multi-sentence semantic phenomena. We thus work towards transforming these QA datasets into NLI datasets with long premises. 
We evaluate models trained on these transformed datasets on two downstream tasks - Multiple-Choice Reading Comprehensions (MCRC) in the QA domain and Checking Factual Correctness of Summaries (CFCS) in the text-summarization domain. Both of these tasks can be reduced to an NLI form (Figure~\ref{fig:intro_fig}).

The main contributions of this paper are as follows: (1) We argue that models trained on existing NLI datasets lack the multi-sentence reasoning capabilities that are needed for downstream tasks such as Question Answering and Summarization. (2) To train NLI models capable of multi-sentence reasoning, we present and analyze three different conversion methods to transform existing MCRC datasets to multi-sentence NLI datasets. We validate the quality of the converted datasets by showing that models trained on them have performance competitive to existing MCRC models\hide{ on  the task of MCRC}.
(3) Our results and analysis show that due to the presence of multi-sentence premises, models trained on the converted NLI datsets perform better than those trained on single-sentence NLI datasets, on both MCRC and CFCS downstream tasks.

%% file: Related Work/related_work.tex
NLI has gained significant attention due to the availability of large scale datasets~\cite{Bowman2015,mnli} that can be used to train data-hungry deep learning models~\cite{kapanipathi2019infusing,wang2015learning}, including transformer-based architectures~\cite{devlin2018bert}.  
However, work relevant to the use of these NLI models for downstream tasks has been very limited and can be categorized into two categories: (1) work focusing on using models trained on sentence-level NLI datasets with fixed or learned aggregation to perform a target downstream task~\cite{falke2019ranking,Trivedi2019}; and (2) work addressing the need for task-specific NLI datasets~\cite{kryscinski2019evaluating,Demszky2018,dnli}.

Recent efforts to apply models trained on sentence-level NLI datasets on downstream NLP tasks such as MCRC and CFCS have had limited success. 
\citet{Trivedi2019} use simple rules to first cast the problem of MCRC to NLI. and subsequently divide the long passage into smaller sentence-level premises. They use a pre-trained NLI model to obtain sentence-level relevance scores with respect to a particular hypothesis combined with a \emph{learned} representation aggregation module to obtain the score for that hypothesis. \citet{falke2019ranking} apply a similar approach for the task of CFCS, and divide both the provided summary as well as the source documents into single-sentence premises and hypotheses. They use a simple entailment score aggregation over all sentence-level premise-hypothesis pairs to obtain the factual correctness score for each provided summary. Both these works note that
models trained on sentence-level NLI datasets do not transfer well to the task of MCRC and CFCS. We argue that this \emph{divide and conquer} approach is not ideal for the problem, and highlight the need for a \emph{native} multi-sentence inference model.

To facilitate the direct \hide{direct is meant to show without finetuning} use of NLI models on downstream tasks like MCRC and CFCS, an interesting alternate approach has been to re-cast datasets from other tasks into NLI datasets. 
\citet{khot2018scitail} use manual annotation to re-cast SciQ (a QA dataset) to SciTail -- an NLI dataset. However, \citet{clark2018think} show that an NLI model trained on SciTail does not perform well on the task of MCRC. 
Similarly, \citet{kryscinski2019evaluating} create an automatically generated training dataset for CFCS. Even though this data has premises consisting of multiple sentences, the analysis done by \citet{FactCCAnalysis} finds that a model trained on this data works well only for summaries having high token overlap with the source.
\citet{Demszky2018} attempt to create an NLI dataset \hide{They create single large dataset} that requires inter-sentence reasoning by converting subsets of various QA datasets to NLI. They try two approaches for the conversion -- rule-based and neural. Their neural approach uses a trained seq2seq BiLSTM-with-copy model ~\cite{gu2016copy} to convert each $\langle${\tt question, answer}$\rangle$ pair into a hypothesis sentence (the corresponding passage being the premise). While their approach looks promising, they do not show the utility of these converted datasets by training an NLI model on them. This makes it unclear whether the NLI datasets generated by the conversion are useful for any downstream task. 
We posit that this direction of research is promising and largely unexplored. Hence, in our work, we attempt to leverage the broad spectrum of MCRC datasets by recasting them to NLI datasets, and show their usefulness by performing the downstream tasks of MCRC and CFCS. 

%% file: NLI Models for Downstream Tasks/nli_for_downstream_NLP.tex
 NLI is usually cast as a multi-class classification problem, where given a premise and a hypothesis, the model classifies the relation between them as \emph{entails}, \emph{contradicts}, or \emph{neutral}. It can also be cast into a two-class problem, where the \emph{contradicts} and \emph{neutral} classes are clubbed into a \emph{not-entails} class. For all our experiments and analysis, we pose NLI as a two-class problem.
We investigate the usefulness of NLI for the downstream tasks of {M}ultiple {C}hoice {R}eading {C}omprehension (MCRC) and  {C}hecking {F}actual {C}orrectness of Text-{S}ummarization (CFCS).
%
 
\vspace{2mm}
\noindent \textbf{MCRC} 
can be cast as an NLI task by viewing the given context as the premise, and the transformed question-answer combinations as different hypotheses~\cite{Trivedi2019}. The multiple answer-option setting can then be approached as: a) individual option entailment tasks, where more than one answer-option can be correct; or b) a multi-way classification task by selecting the answer-option which gets the highest entailment score from the model, when only a single correct answer-option exists.

\vspace{2mm}
\noindent \textbf{CFCS} 
 can also be reduced to a two-class NLI problem. A factually correct summary should be entailed by the given source text -- it should not contain {\em hallucinated facts},  
and it should also not contradict facts present in the source text. 

\vspace{1mm}
\noindent 
Despite being ideally suited for reduction to NLI, both MCRC and CFCS have proved to be difficult to solve using models trained on single-sentence NLI datasets~\cite{Trivedi2019,falke2019ranking}. 


%% file: NLI Models for Downstream Tasks/long_premise_conjecture.tex
\begin{table}[h!]
\centering
\resizebox{0.49\textwidth}{!}{%
\begin{tabular}{
    l|
    l|
    c|
    c}
\toprule
    \textbf{Task} 
    & \textbf{Dataset} 
    & \textbf{
        \begin{tabular}
            [c]{@{}c@{}}
            Word\\ Count \\ (Avg)
        \end{tabular}
        }
    & \textbf{
        \begin{tabular}
            [c]{@{}c@{}}
            Sentence\\ Count \\ (Avg)
        \end{tabular} 
        }
    \\ 
\midrule
    \multicolumn{1}{l|}{\multirow{2}{*}{NLI}}  
        & \multicolumn{1}{l|}{MultiNLI} 
            & 22  
            & 1.1
        \\ 
    \multicolumn{1}{l|}{}  
        & \multicolumn{1}{l|}{SNLI}     
            & 14  
            & 1.0
        \\ 
\midrule
    \multicolumn{1}{l|}{\multirow{4}{*}{MCRC}} 
        & \multicolumn{1}{l|}{RACE}
            & 271 
            & 18.5
        \\
    \multicolumn{1}{l|}{}
        & \multicolumn{1}{l|}{MultiRC}  
            & 252 
            & 14.3
        \\
    \multicolumn{1}{l|}{}
        & \multicolumn{1}{l|}{DREAM}
            & 110 
            & 13.9
        \\
    \multicolumn{1}{l|}{}
        & \multicolumn{1}{l|}{CosmosQA} 
            & 75  
            & 3.8
        \\
\midrule
    \multicolumn{1}{l|}{\multirow{2}{*}{CFCS}} 
        & \multicolumn{1}{l|}{FactCC}
            & 546 
            & 28.5
        \\ 
    \multicolumn{1}{l|}{}
        & Summary Reranking
            & 738 
            & 29.5
        \\ 
\bottomrule
\end{tabular}%
}
\caption{The average premise length in various datasets. The key point to notice here is the sharp increase in premise lengths from NLI datasets to MCRC and CFCS datasets.}
\label{tab:premise_lengths}
\end{table}

Datasets for the downstream MCRC and CFCS tasks contain significantly longer \hide{premise}texts than the single-sentence NLI datasets (Table \ref{tab:premise_lengths}). 
This shift in the \hide{premise}text length brings about a fundamental change in the nature of the NLI problem. Performing inference over longer forms of text requires a multitude of additional reasoning skills like coreference resolution, event detection, dialogue understanding, abductive reasoning etc. ~\cite{cooper1996usingsemantic,lai-etal-2017-natural,Demszky2018}.
These are over and above the reasoning types needed to perform inference locally at sentence level. Thus, models trained on sentence-level NLI datasets are incapable of performing multi-sentence inference, which we posit as the main cause for their low performance on downstream tasks like CFCS and MCRC. 



\begin{table}[]
\centering

\begin{tabular}{
    l|
    l|
    c}
\toprule
    \textbf{Task} 
    & \textbf{Dataset} 
     & \textbf{Dataset Size} 
    \\ 
\midrule
    \multicolumn{1}{l|}{\multirow{4}{*}{MCRC}} 
        & \multicolumn{1}{l|}{RACE}
             & 87866 
        \\
    \multicolumn{1}{l|}{}
        & \multicolumn{1}{l|}{MultiRC}  
             & 27243 
        \\
    \multicolumn{1}{l|}{}
        & \multicolumn{1}{l|}{DREAM}
             & 6116
        \\
    \multicolumn{1}{l|}{}
        & \multicolumn{1}{l|}{CosmosQA} 
             & 23766 
        \\
\midrule
    \multicolumn{1}{l|}{\multirow{2}{*}{CFCS}} 
        & \multicolumn{1}{l|}{FactCC}
             & 931
        \\ 
    \multicolumn{1}{l|}{}
        & Summary Reranking
             & 1000
        \\ 
\bottomrule
\end{tabular}%

\caption{The number of annotated instances in MCRC and CFCS datasets. MCRC is an extremely resource-rich task whereas CFCS is considerably resource-deficient.
}
\label{tab:dataset_size}
\end{table}

In order to train models capable of performing multi-sentence inference, we need NLI datasets that possess longer multi-sentence premises.
The challenge, however, is to obtain such  datasets.
The paucity of multi-sentence NLI datasets can be overcome by transforming large MCRC datasets into NLI datasets through a quality preserving transformation procedure. While the task of CFCS also provides a similar opportunity, the sheer lack of annotated training instances inhibits its use. Table~\ref{tab:dataset_size} shows the abundance of training instances in MCRC datasets, and highlights the deficiency in CFCS datasets.
Hence, in this work, we rely on various MCRC datasets to provide this data. 

 In the following section, we present  three methods to reformat MCRC datasets to create multi-sentence NLI datasets. 
 We then evaluate models trained on these multi-sentence NLI datasets on the tasks of MCRC and CFCS, and contrast their performance with those trained on a single-sentence NLI dataset. 
 

%% file: Reformatting QA to NLI/reformatting_qa_to_nli.tex
\begin{table*}[!ht]
\small
    \centering
    \begin{tabular}{p{0.23\textwidth}|p{0.23\textwidth}|p{0.23
    \textwidth}|p{0.23\textwidth}}
    \toprule
        & \textbf{Rule-based} 
        & \textbf{Neural} 
        & \textbf{Hybrid} \\
    \midrule
        \textbf{Q:} What building were the four captives inside on Tuesday?
            \break \textbf{A:} CNN headquarters
            & The four captives inside on Tuesday were CNN headquarters.
            & The four captives were inside CNN headquarters on Tuesday. 
            & The four captives were inside CNN headquarters on Tuesday.\\
            \hline
        \textbf{Q:} How do suburban commuters travel to and from the city in Copenhagen at present?
            \break \textbf{A:} About one third of the suburban commuters travel by bike.
            & Suburban commuters travel to about one third of the suburban commuters travel by bike and from the city in Copenhagen at present.
            & Suburban commuters travel to and from the city in Copenhagen at present by bike 
            & Suburban commuters travel to about one third of the suburban commuters travel by bike and from the city in Copenhagen at present.\\
    \bottomrule
    \end{tabular}
    \caption{Examples of Rule-based, Neural and Hybrid Conversions}
    \label{tab:example_conversions}
\end{table*}

As shown in Figure \ref{fig:intro_fig}, we can convert MCRC datasets into two-class NLI datasets by reusing the passage as a premise and paraphrasing the question along with each answer option as individual hypothesis options.
The following describes the different conversion techniques we use for this.

%% file: Reformatting QA to NLI/rule.tex
In the rule-based method of conversion, we use the Stanford CoreNLP package \cite{stanfordnlp} to generate the dependency parse of both the question and the answer option, followed by the application of conversion rules proposed by  \citet{Demszky2018} to generate a hypothesis sentence. 
However, due to the limited coverage of rules and errors in the dependency parse, some of the generated hypotheses sound unnatural (first example in Table \ref{tab:example_conversions}). In order to generate more natural and diverse hypotheses and to get broader coverage in conversion, we implement a neural conversion strategy.

%% file: Reformatting QA to NLI/neural.tex
Due to the recent success of transformer-based text generation models, we train a BART \cite{bart} model to generate a grammatically coherent hypothesis from question + answer option (word/phrase) as input. We use a sequence of datasets as a curriculum to finetune the BART conversion model: (1) starting with CNN/Daily Mail summarization dataset \cite{cnn}, which makes the generated sentences coherent; (2) followed by Google's sentence compression dataset \cite{gsc}, which limits the generated sequence to a single sentence; and (3) finally the annotated dataset provided by \citet{Demszky2018}\footnote{We refer to the annotated dataset provided by \citet{Demszky2018} as QA2D.} which has around $71000$ (question-answer, hypothesis) pairs 
from various QA datasets. Based on manual inspection, we find that the hypotheses generated by this method indeed sound more natural and diverse than the ones produced by the rule-based conversion\footnote{More examples of conversion results are presented in Appendix \ref{app:conversion}.}.
In some cases, however, the generated hypotheses either discard some crucial information, or contain hallucinated facts that do not convey the exact information in the source question-answer pair (Table \ref{tab:example_conversions}).
We thus define a hybrid conversion strategy, combining the desirable aspects of the rule-based conversion and the neural conversion strategies.

%% file: Reformatting QA to NLI/hybrid.tex

We design a heuristic to compose a hybrid dataset to overcome the caveats in the neural conversion. We use the number of words in the question-answer concatenation as a proxy for the expected length of the hypothesis. We target the problems of hallucination and missing information in the neural conversions by accepting only those neural-generated hypotheses that lie in the range of $0.8$ and $1.2$ times the length of the question-answer concatenation. We replace the rejected neural hypotheses with the rule-based hypothesis, if rule-based conversion is feasible; or with the question-answer concatenation otherwise; as seen in Table \ref{tab:example_conversions}. 
The selection policy is driven by the need to get more natural and coherent conversions without compromising on the accuracy and preservation of factual information in the question and answer option. The choice of the specific range is purely empirical in nature.
In Section \ref{sec:results_and_discussion}, we discuss in detail the effectiveness of the heuristically combined dataset.

%% file: A Transferable NLI Model/transferable_modeling.tex
\begin{figure*}[!ht]
    \centering
    \begin{minipage}{0.4\textwidth}
        \centering
        \includegraphics[width=0.9\textwidth]{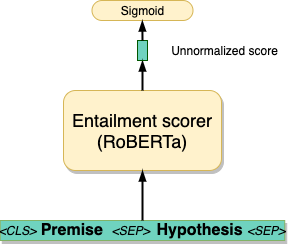} 
        \caption{Two class entaiment model.}
        \label{fig:model_entailment}
    \end{minipage}\hfill
    \begin{minipage}{0.45\textwidth}
        \centering
        \includegraphics[width=0.9\textwidth]{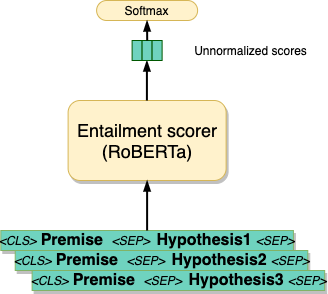} 
        \caption{Multiple choice classification model.}
        \label{fig:model_mc}
    \end{minipage}
    \label{fig:model}
\end{figure*}

In order to use pretrained NLI models for the tasks of MCRC and CFCS, we need 
the model
to be agnostic to the peculiarities of the downstream task.
This can be achieved by dividing the model architecture into two parts : (1) a transferable entailment scorer; and (2) a weight-free comparator on top of that scorer. Each premise-hypothesis pair is encoded as a single sequence pair and passed through the transferable entailment scorer to produce an entailment score. Depending on the problem setup, the comparator can either be a sigmoid function (for a two-class entailment problem) as shown in Figure~\ref{fig:model_entailment}; or a softmax function (for multiple choice classification) as shown in Figure~\ref{fig:model_mc}. This logical segmentation of the model makes it easy to transfer the model weights across different tasks. For the entailment scorer, we use a 2-layer feed-forward network on top of the [CLS] token of pretrained RoBERTa \footnote{The RoBERTa model is pretrained on the masked language modelling objective as described in \citet{liu2019roberta}. We obtain it from the HuggingFace library ~\cite{huggingfaceTransformers}.}.

In our experiments evaluating the transferability of the entailment model, we perform various zero-shot evaluations. This requires interpreting the entailment scores a bit differently for each task. To transfer the weights from a multiple choice classification model (Figure~\ref{fig:model_mc}) to a two class entailment model (Figure~\ref{fig:model_entailment}), we copy the weights of the transferable entailment scorer as-is, and calibrate a threshold using a dev set to interpret the outputs from the sigmoid comparator for binary classification. 
Since the softmax comparator does not need any calibration, the transfer in the other direction, i.e., from a two class entailment model to a multiple choice classification model is more straightforward -- we simply copy the weights of the transferable entailment scorer. 

%% file: Datasets/datasets.tex
For our experiments, we use 4 MCRC datasets and their transformed NLI versions; 2 CFCS datasets; and 1 single-sentence NLI dataset. These datasets are qualitatively described below: 
    
    \vspace{2mm}
    
    \noindent\textbf{Single-sentence NLI Dataset:} 
    
    \vspace{0.5mm}
    
    \noindent\textbf{MultiNLI}~\cite{mnli} is chosen as the single-sentence NLI dataset as it is widely used to learn and evaluate sentence-level NLI models.
    
    \vspace{2mm}
   
    \noindent \textbf{MCRC Datasets: }
    
        \vspace{0.5mm}
        
        \noindent \textbf{RACE}~\cite{lai2017race} broadly covers  detail reasoning, whole-picture reasoning, passage summarization, and attitude analysis.
        
        \vspace{0.5mm} 
        
        \noindent\textbf{MultiRC}~\cite{multirc} mainly contains questions which require multi-hop reasoning and coreference resolution.
        
        \vspace{0.5mm}
        
        \noindent\textbf{DREAM}~\cite{sun2019dream} is a dialogue-based MCRC dataset, where the context is a multi-turn, multi-party dialogue. 
        
        \vspace{0.5mm}
        
        \noindent\textbf{CosmosQA}~\cite{huang2019cosmos} focuses more on commonsense and inductive reasoning, which require reading between the lines.\footnote{Questions where the answer is ``None of the above'' are removed from the CosmosQA dataset.}
    
    \vspace{2mm}
    
    \noindent \textbf{CFCS Datasets: }
    
    \vspace{0.5mm}
    
        \noindent\textbf{FactCC \hide{(evaluation set)}}~\cite{kryscinski2019evaluating}
     consists of tuples of the form $\langle${\tt article, sentence}$\rangle$, where the articles are taken from the CNN/DailyMail corpus, and sentences come from the summaries for these articles generated using several state-of-the-art abstractive summarization models. 
     
    \vspace{0.5mm}
    
       \noindent\textbf{Ranking Summaries for Correctness (evaluation set)}~\cite{falke2019ranking} 
    consists of articles and a set of summary alternatives for each article, where some of the provided summaries are factually inconsistent w.r.t the article.

%% file: Results and Discussion/results_and_discussion.tex
In this section, we discuss the quality of the converted datasets, and the ability of models trained on these datasets to transfer knowledge to the downstream tasks of MCRC and CFCS. We contrast the performance of these models with a model trained on MultiNLI to assert the utility of the converted datasets on long premise downstream tasks; and in this process evaluate the long premise conjecture.

%% file: Results and Discussion/qa_vs_nli.tex
In order to evaluate the quality of conversion, we compare the NLI models trained on the converted datasets to their corresponding MCRC QA models. 
For this, we finetune RoBERTa in the multiple-choice classification setting (Figure \ref{fig:model_mc}) on each of the converted datasets. In order to set the performance bar, we also train RoBERTa Q+A concatenation models\footnote{Q+A concatenation form follows the work of \citet{liu2019roberta} and sets a very strong quality bar.} on each of the corresponding MCRC datasets.

\begin{table}[!ht]
\centering
\begin{tabular}{@{}lccc@{}}
\toprule
    \multirow{2}{*}{\textbf{Dataset}} 
    & \multicolumn{3}{c}{\textbf{\begin{tabular}[c]{@{}c@{}}Dataset Format \\ (conversion method)\end{tabular}}} \\ \cmidrule(l){2-4} 
    & \textbf{\begin{tabular}[c]{@{}c@{}}QA \\ (Q+A)\end{tabular}} 
    & \textbf{\begin{tabular}[c]{@{}c@{}}NLI \\ (Neural)\end{tabular}} 
    & \textbf{\begin{tabular}[c]{@{}c@{}}NLI \\ (Hybrid)\end{tabular}} \\
\midrule
    RACE
        & 84.33 
        & 82.89 
        & 83.99 \\
    DREAM
        & 84.22
        & 82.41 
        & 83.29 \\
    MultiRC 
        & 85.19
        & 80.60
        & 81.22\\
    CosmosQA
        & 85.58
        & 83.34 
        & 83.89 \\
\bottomrule
\end{tabular}
\caption{Test set accuracy of models trained on converted forms of different MCRC datasets, formed using the three conversion strategies described in Section \ref{sec:conversion}.}
\label{tab:all-results}
\end{table}

Table \ref{tab:all-results} shows that models trained on the converted datasets achieve performance
comparable to the corresponding Q+A models for each of the four MCRC datasets. 
From this result, we can infer that the conversion mechanism captures most of the information from the MCRC datasets. Further, the models trained on the datasets formed by the hybrid conversion strategy perform consistently better in comparison to their pure neural counterparts.
This substantiates the motivation for performing the hybrid conversion strategy as discussed
in Section \ref{subsec:conversion_hybrid}, and shows that the resulting hybrid conversion approach does produce better quality conversions than neural or rule-based alone.
Hence, we only use the NLI dataset obtained using the hybrid conversion technique for our analysis in the subsequent experiments.

Having ascertained the quality of the converted NLI datasets, we discuss the experiments performed to substantiate our long premise conjecture by performing the task of MCRC and CFCS.

%% file: Results and Discussion/LPC.tex

To validate the long premise conjecture, we perform the tasks of MCRC and CFCS using pretrained NLI models. Specifically, we analyze and contrast the performance of NLI models trained on the sentence-level NLI dataset -  MultiNLI, with those trained on multi-sentence NLI datasets obtained by converting the four MCRC datasets using the hybrid conversion strategy described in Section \ref{sec:conversion}.
We compare the zero-shot performance of these models on the MCRC and CFCS datasets described in Section \ref{sec:diverse_qa_datasets}. All MCRC evaluations are performed using the transformed NLI version of the data. 
Since MultiNLI is a single-sentence NLI dataset, the model trained on MultiNLI is evaluated in two ways: (1) by passing the entire premise; and (2) by segmenting the premise into individual sentences and aggregating the entailment score with respect to all the segments.

%% file: Results and Discussion/LPC_MCRC.tex
We evaluate each of the five models on all the MCRC datasets (in NLI form) and discuss the performances here.
The results of these evaluations are presented in Table \ref{tab:results_long_premise}.  We show that, in most cases, the models trained on the converted NLI datasets outperform the MultiNLI model on all \textit{other}
\footnote{Other refers to all the MCRC datasets (shown in Section \ref{sec:diverse_qa_datasets}) except the one on which the model is trained.} MCRC datasets.
We assert that this difference in performance can be attributed to the difference in premise lengths of the converted datasets and MultiNLI.

\begin{table}[]
    \centering
    \resizebox{\columnwidth}{!}{%
    \begin{threeparttable}
    \begin{tabular}{l|c|c|c|c}
    \toprule
        \multirow{2}{*}{\diagbox[]{\textbf{Model}}{\textbf{Dataset}\tnote{\dag}}}
            & RACE
            & MultiRC
            & DREAM
            & CosmosQA\\
            & (271)
            & (252)
            & (110)
            & (75)\\
    \midrule
        {\small Random Guess}
            & 25.00
            & 50.00
            & 33.33
            & 33.33\\
    \midrule
        MultiNLI
            & 44.34
            & 60.58
            & 67.76
            & 38.11\\
        MultiNLI\textsubscript{Segmented}
            & 41.01
            & 61.71
            & 42.28
            & 43.28\\
        RACE\textsubscript{Hybrid}
            & \hide{\textit{83.99}}\tnote{x}
            & \textbf{77.43}
            & \textbf{83.58}
            & \textbf{73.58}\\
        MultiRC\textsubscript{Hybrid}
            & 58.02
            & \hide{\textit{81.22}}\tnote{x}
            & 67.12
            & 43.65\\
        DREAM\textsubscript{Hybrid}
            & \textbf{65.01}
            & 71.08
            & \hide{\textit{83.99}}\tnote{x}
            & 61.00\\
        CosmosQA\textsubscript{Hybrid}
            & 49.27
            & 48.80
            & 72.46
            & \hide{\textit{83.89}}\tnote{x}\\
    \bottomrule
    \end{tabular}
    \begin{tablenotes}[flushleft]
    {
    \item [\dag] Datasets are in NLI form created using hybrid conversion method (Section \ref{subsec:conversion_hybrid}) for the hybrid models 
    \item [x] These numbers are not presented as they are not the result of zero-shot evaluation. 
    Refer Table \ref{tab:all-results} for them.}
    \end{tablenotes}
    \end{threeparttable}
    }
    \vspace{-1.8mm}
    \caption{Zero-shot evaluation accuracies achieved by models trained on converted NLI datasets and MultiNLI on \emph{other} MCRC datasets (in NLI form) using the transferable model architecture described in Section \ref{sec:transferable_modeling}. The numbers in the parenthesis of the column headers denote the average premise lengths of the datasets.}
    \label{tab:results_long_premise}
    
\end{table}

\begin{figure}
    \centering
    \hspace{-10pt}\includegraphics[width=0.5\textwidth]{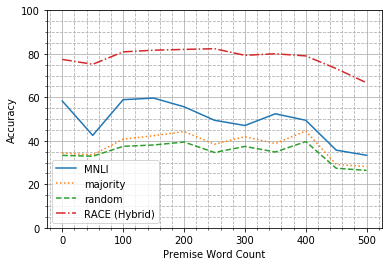}
    \caption{The graphs show the performance of models trained on RACE\textsubscript{Hyrbid} and MultiNLI at different premise lengths on a combined evaluation set of all the MCRC datasets mentioned in Section \ref{sec:diverse_qa_datasets}.
    . 
    The accuracy at length {\em x} denotes the accuracy of the models on the examples with premise length in $[x, x+50)$ words.}
    \label{fig:long_premise_conjecture_mcrc}
\end{figure}

To present further evidence in support of our claim, we analyse the performance of the models trained on the hybrid conversion of RACE (RACE\textsubscript{Hybrid})
and MultiNLI, with varying premise length. For the purpose of this experiment, we combine all the MCRC dev datasets described in Section \ref{sec:diverse_qa_datasets} into a single large dev set, and plot the performance with respect to the number of words in the premise in buckets of size 50. Figure \ref{fig:long_premise_conjecture_mcrc} shows the sharp decline in the performance of the MultiNLI model as the length of the premise increases beyond 150 words, whereas RACE\textsubscript{Hybrid} is much more robust to increases in premise length.

%% file: Results and Discussion/LPC_CFCS.tex
\hide{
\newlyadded{To further defend
our long-premise conjecture
and show the utility of long-premise NLI datasets,
we use}
them to solve the resource deficient task of checking factual consistencies of summaries (CFCS). The task can be setup the following two forms: 
}

We show the utility of long-premise NLI datasets by performing the CFCS task, which can be set up in the following two ways:

\vspace{2mm}
\noindent\textbf{(1) CFCS as classification}
\vspace{1mm}

\noindent In this form, given a document and a corresponding summary sentence, the model needs to identify if the sentence is factually correct with respect to the document (is entailed) or not. 
In order to perform the classification, we first obtain our entailment scorer by fine-tuning the multiple choice classification model (Figure \ref{fig:model_mc}) on the NLI form of the RACE dataset and use the dev set\footnote{\label{fn:factcc}We use the dev and test dataset provided by \citet{kryscinski2019evaluating} for this task.} to calibrate a threshold\footnote{Balanced accuracy is used to find the best threshold.} as described in Section \ref{sec:transferable_modeling} to obtain the two-class entailment model (Figure \ref{fig:model_entailment}).

\vspace{2mm}

\noindent\textbf{(2) CFCS as ranking}

\vspace{1mm}

\noindent Given a source document and a set of five\footnote{A variable number of these five machine generated summaries can be factually correct. However, there is always at least one incorrect summary in this set.} machine generated summaries, the model is required to rank at least one factually correct summary above all incorrect summary alternatives. 
We also solve the auxiliary diagnostic task of sentence-pair ranking, where the premise document $D=\{d_1, d_2, \dots\}$ as well as the summary $S=\{s_1, s_2, \dots\}$ are divided into individual sentences and the model is required to decide if $d_i$ entails $s_j$ or not. 

\begin{table}[!ht]
\centering
\resizebox{\columnwidth}{!}{%
\begin{threeparttable}

\begin{tabular}{@{}l|c|c@{}}
\toprule
    \textbf{Model} 
    & \begin{tabular}[c]{@{}c@{}}\textbf{Balanced}\\ \textbf{Accuracy }\end{tabular} 
    & \textbf{F1-score} \\ 
\midrule
    BERT+FactCC\textsubscript{autogen}\tnote{\textasteriskcentered} ~ \tnote{\#}
        & 74.15 
        & 0.51\\
    RoBERTa 
        & \hide{49.89} 54.76
        & \hide{0.46} 0.30 \\
    RoBERTa+MultiNLI 
        & \hide{50.00} 51.92
        & \hide{0.47} 0.15 \\
    RoBERTa+MultiNLI\textsubscript{Segmented}
        & 69.87
        & 0.70 \\
    RoBERTa+CosmosQA\textsubscript{Hyrbid}
        & \hide{50.00} 55.96
        & \hide{0.47} 0.52\\
    RoBERTa+DREAM\textsubscript{Hyrbid}
        & \hide{66.06} 75.69
        & \hide{0.68} 0.69\\ 
    RoBERTa+MultiRC\textsubscript{Hyrbid}
        & \hide{78.27} 82.03
        & \hide{0.79} {0.72}\\
    RoBERTa+RACE\textsubscript{Hyrbid}
        & \hide{82.99} \textbf{86.55}
        & \hide{0.82} \textbf{0.73}\\ 
\bottomrule
\end{tabular}
\begin{tablenotes}[flushleft]
    \footnotesize
    \item [\textasteriskcentered] These results are reported from~\citet{kryscinski2019evaluating}. 
    \item [\#] FactCC\textsubscript{autogen} is the automatically generated training data used by \citet{kryscinski2019evaluating}.
\end{tablenotes}
\vspace{-5pt}
\caption{Balanced accuracy and macro F1 score on the test set
for the task of CFCS posed as a classification problem.}
\label{tab:results-factcc}
\end{threeparttable}}
\end{table}

\begin{table}[!ht]
\centering
\resizebox{\columnwidth}{!}{%
\begin{threeparttable}

\begin{tabular}{@{}l|c|c@{}}
\toprule
    \multirow{3}{*}{\textbf{Model}} 
        & \multicolumn{2}{c}{\textbf{\% Correct \hide{($\uparrow$)}}} \\
        \cmidrule(lr){2-3} 
        & \begin{tabular}[c]{@{}c@{}}\textbf{Sentence-pair}\\ \textbf{Ranking }\end{tabular} 
        & \begin{tabular}[c]{@{}c@{}}\textbf{Summary}\\ \textbf{Ranking}\end{tabular} \\ 
\midrule
    ESIM + SNLI \tnote{\textasteriskcentered}
        & 67.60\%
        & 60.70\% \\
    BERT+FactCC\textsubscript{autogen} \tnote{\textdagger} ~ \tnote{\#}
            & 70.00\% 
            & - \\
    QAGS\tnote{\ddag}
        & 72.10\% 
        & - \\
    RoBERTa 
        & 56.03\%
        & 50.47\%\\
    RoBERTa+MultiNLI 
        & 81.76\%
        & 49.53\%\\
    RoBERTa+MultiNLI\textsubscript{Segmented}
        & 81.23\%
        & 66.36\%\\
    RoBERTa+CosmosQA\textsubscript{Hyrbid}
        & 76.41\%
        & 49.53\%\\
    RoBERTa+DREAM\textsubscript{Hyrbid}
        & 78.28\%
        & 68.22\%\\ 
    RoBERTa+MultiRC\textsubscript{Hyrbid}
        & 72.21\%
        & 67.23\%\\
    RoBERTa+RACE\textsubscript{Hyrbid} 
        & \textbf{86.59\%}
        & \textbf{75.70\%}\\ 
\bottomrule
\end{tabular}
\begin{tablenotes}[flushleft]
    \item [\textasteriskcentered ~ \textdagger ~ \ddag] Reported from~\citet{falke2019ranking}, ~\citet{kryscinski2019evaluating} and \citet{wang2020asking}, respectively. 
    \item [\#] FactCC\textsubscript{autogen} is the automatically generated training data for their model.
\end{tablenotes}
\end{threeparttable}
}
\vspace{-2mm}
\caption{Performance of various models on the CFCS on the sentence-ranking and summary-ranking tasks. The numbers denote the fraction of highest ranked summaries which are labelled factually correct. 
}
\vspace{-3mm}
\label{tab:results-falke}
\end{table}

\vspace{1mm}

\noindent Table~\ref{tab:results-factcc} and Table~\ref{tab:results-falke} present the results for CFCS as classification and CFCS as ranking, respectively. 
As can be seen, the model performances steadily increase as the premise lengths in the training data increase. 
The model trained on RACE\textsubscript{Hybrid} which has the longest average premise length (c.f. Table \ref{tab:premise_lengths}), outperforms all the models trained on datasets having comparatively shorter premises.
Moreover, it also outperforms the FactCC model which uses the automatically generated long-premise training data \cite{kryscinski2019evaluating}.
Another insightful observation is that the model trained on MultiNLI performs the short-premise task of sentence-pair ranking reasonably well, but is unable to translate this to the task of summary ranking (which has long premises).

\begin{figure}[!ht]
    \centering
    \hspace{-10pt}
    \includegraphics[width=\columnwidth]{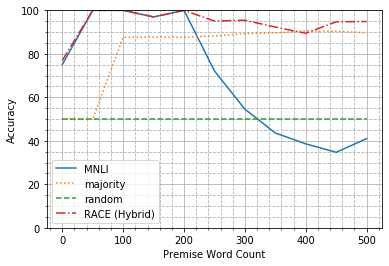}
    \caption{The graphs show the performance of models trained on RACE\textsubscript{Hyrbid} and MultiNLI at different premise lengths on a combined evaluation set of all the CFCS datasets mentioned in Section \ref{sec:diverse_qa_datasets}.
    . 
    The accuracy at length {\em x} denotes the accuracy of the models on the examples with premise length in $[x, x+50)$ words.}
    \label{fig:long_premise_conjecture_cfcs}
\end{figure}

We also repeat the experiment of evaluating the models on examples with varying premise lengths by using a combination of all the CFCS dev datasets described in Section \ref{sec:diverse_qa_datasets}. Figure \ref{fig:long_premise_conjecture_cfcs} shows the considerably steeper decline in the performance of the MultiNLI model as compared to the RACE\textsubscript{Hybrid} model as the premise length increases beyond $200$, similar to the trend observed on the task of MCRC (Figure \ref{fig:long_premise_conjecture_mcrc}).

The results of evaluations on both the downstream tasks provide sufficient evidence supporting our long premise conjecture. Moreover, since the models trained on the converted data outperform all the reported results on the task of CFCS, they can act as a rich resource for the community for this task.

%% file: Conclusion/conclusion.tex
%
%
The difficulty of transferring entailment knowledge to downstream NLP tasks can be largely attributed to the difference in data distributions, specifically the premise lengths.
Models trained on single-sentence NLI datasets are incapable of performing the multi-sentence inference required for the downstream tasks.
%
%
 
 We experiment with three conversion strategies -- rule-based, neural, and hybrid -- to recast existing MCRC datasets into NLI datasets. We discuss the trade-off between structure and grammatical coherence in the context of the conversion, and perform experiments to identify the hybrid conversion strategy as the best.
%
%
 Recasting MCRC datasets into NLI using this strategy can result in  broadly useful NLI datasets. 
 Models trained on these multi-sentence NLI datasets perform
 better than models trained on existing single-sentence NLI datasets,
 in the context of the long-premise downstream tasks of MCRC and CFCS. These datasets can be a useful resource for creating truly generalizable NLI models.


%% file: appendix/appendices.tex
\setcounter{footnote}{0}
\section{Reproducibility Checklist}
\label{app:reproducibility}
\subsection{Details of the datasets used}
\label{app:datasets}
\input{appendix/datasets}

\subsection{Neural Conversion}
We use the following training sequence to obtain the final neural conversion model:
\begin{enumerate}
    \item Obtain the pre-trained BART model \cite{bart} fine-tuned on CNN/Dailymail from HuggingFace  library.\footnote{https://huggingface.co/facebook/bart-large-cnn}
    \item Fine-tune the model using the hyperparameters mentioned in Table \ref{tab:hyp_neural_conversion} on google-sentence completion dataset \cite{gsc}\footnote{\url{https://github.com/google-research-datasets/sentence-compression}}
    \item Further fine-tune the model on the QA2D datatset \cite{Demszky2018}.\footnote{\url{https://worksheets.codalab.org/worksheets/0xd4ebc52cebb84130a07cbfe81597aaf0/}}
\end{enumerate}

\begin{table}[!ht]
\centering
\resizebox{\columnwidth}{!}{%
\begin{tabular}{@{}lcc@{}}
\toprule
\textbf{Hyperparam}      & \multicolumn{2}{c}{\textbf{Dataset/fine-tune curriculum step}}                               \\ \cmidrule{2-3}
                         & \begin{tabular}[c]{@{}c@{}}Google-sentence \\ compression\end{tabular} & QA2D                \\\midrule
learning rate            & 1e-5                                                                   & 1e-5                \\
weight decay             & 0.01                                                                   & 0.01                \\
adam epsilon             & 1e-8                                                                   & 1e-8                \\
max. grad. norm          & 1.0                                                                    & 1.0                 \\
warmup steps             & 1125                                                                   & 600                 \\
batch size               & 24                                                                     & 32                  \\
max epochs               & 3                                                                      & 5                   \\
max seq. len             & 50                                                                     & 50                  \\
lower-case               & False                                                                  & False               \\ \midrule
\textbf{Runtime metrics} &                                                                        &                     \\ \midrule
Python                   & 3.7.4                                                                  & 3.7.4               \\
GPU Type                 & GeForce RTX 2080 Ti                                                    & GeForce RTX 2080 Ti \\
Num. GPUs                & 1                                                                      & 1                   \\ \bottomrule
\end{tabular}%
}
\caption{Hyperparameters and runtime metrics for training the neural conversion model}
\label{tab:hyp_neural_conversion}
\end{table}

\subsubsection{Experiments}
\begin{itemize}
    \item The hyperparams for the models used throughout the Section \ref{sec:results_and_discussion} are shown in Table \ref{tab:hyperparams_models}. These were obtained using minimal manual tuning.
    
    \item The threshold for CFCS as classification experiments (Section \ref{subsubsec:results_cfcs} (1)) we calculated by tuning for best balanced accuaracy \url{https://scikit-learn.org/stable/modules/generated/sklearn.metrics.balanced_accuracy_score.html}. 
\end{itemize}
\begin{table*}[]
\centering
\resizebox{\textwidth}{!}{%
\begin{tabular}{l|lcccl}
                         & \multicolumn{5}{c}{\textbf{Model}}                                                                                                             \\ \hline
\textbf{Hyperparam}      & \textbf{\textbf{RoBERTa+RACE}} & \textbf{RoBERTa+DREAM}    & \textbf{RoBERTa+MultiRC}  & \textbf{RoBERTa+CosmosQA} & \textbf{RoBERTa+MultiNLI} \\ \hline
learning rate            & 1e-5                           & 1e-5                      & 1e-5                      & 1e-5                      & 1e-5                      \\
weight decay             & 0.001                          & 0.1                       & 0.001                     & 0.1                       & 0.01                      \\
max. grad. norm.         & 1.0                            & 1.0                       & 1.0                       & 1.0                       & 1.0                       \\
warmup steps             & 1300                           & 500                       & 300                       & 500                       & 1200                      \\
batch size               & 24                             & 32                        & 32                        & 24                        & 48                        \\
max epochs               & 4                              & 10                        & 4                         & 4                         & 4                         \\ \hline
\textbf{Runtime metrics} &                                & \multicolumn{1}{l}{}      & \multicolumn{1}{l}{}      & \multicolumn{1}{l}{}      &                           \\ \hline
Python                   & 3.7.3                          & \multicolumn{1}{l}{3.7.3} & \multicolumn{1}{l}{3.7.3} & \multicolumn{1}{l}{3.7.3} & 3.7.3                     \\
GPU type                 & m40                            & \multicolumn{1}{l}{m40}   & \multicolumn{1}{l}{m40}   & \multicolumn{1}{l}{m40}   & Titan X                   \\
Num. GPUs                & 1                              & \multicolumn{1}{l}{1}     & \multicolumn{1}{l}{1}     & \multicolumn{1}{l}{1}     & 1                         \\ \hline
Final dev accuracy &
  \begin{tabular}[c]{@{}l@{}}83.08 (Q+A)\\ 82.02(Neural)\\ 84.00(Hybrid)\end{tabular} &
  \multicolumn{1}{l}{\begin{tabular}[c]{@{}l@{}}84.36 (Q+A)\\ 84.07 (Neural)\\ 84.12 (Hybrid)\end{tabular}} &
  \multicolumn{1}{l}{\begin{tabular}[c]{@{}l@{}}84.28 (Q+A)\\ 80.16(Neural)\\ 79.94 (Hybrid)\end{tabular}} &
  \multicolumn{1}{l}{\begin{tabular}[c]{@{}l@{}}85.33 (Q+A)\\ 83.65 (Neural)\\ 83.91 (Hybrid)\end{tabular}} &
  93.44 \\ \hline
\end{tabular}%
}
\caption{Hyperparam setting for the models trained on MCRC datasets and MultiNLI (same for Q+A, Neural, and Hybrid from). These are common for all models in the experiments (Section \ref{sec:results_and_discussion}).}
\label{tab:hyperparams_models}
\end{table*}

\section{Conversion examples}
\label{app:conversion}
\input{appendix/conversion_examples}

%% file: appendix/datasets.tex
Table \ref{tab:datasets_split} gives the train/dev/test splits of the various source datasets used in this work. We follow the same splits after the conversion to NLI form. Since the test datasets are not openly available for MultiRC and CosmosQA,  we use the corresponding dev sets to report our results.
\begin{table}[!ht]
\centering
\begin{tabular}{@{}lccc@{}}
\toprule
\multirow{2}{*}{Dataset}  & \multicolumn{3}{c}{Number of examples} \\\cmidrule(l){2-4} 
               & Train   & Dev  & Test \\
\bottomrule
RACE           & 87866   & 4887 & 4934 \\
MultiRC        & 27243   & 4848 & - \\
DREAM          & 6116    & 2040 & 2041 \\
CosmosQA       & 6116    & 2040 & - \\
FactCC         & -       & 931  & 503 \\
Sentence Ranking     & - & 746  & - \\
Summary Ranking      & - & 2555 & 530 \\\bottomrule
\end{tabular}
\caption{Number of examples in each of the datasets.}
\label{tab:datasets_split}
\end{table}

Table \ref{tab:hybrid_splits} shows the proportion (absolute numbers) of neural, rule-based and Q+A examples in the final hybrid datasets. 

\begin{table}[]
\resizebox{\columnwidth}{!}{%
    \centering
    \begin{tabular}{l|l|c|c|c}
    \toprule
        \textbf{Dataset} & \textbf{Split} & \textbf{Neural} & \textbf{Rule-based} & \textbf{Q+A}\\
    \midrule
        \multirow{3}{*}{RACE} 
            & Train & 314448 & 16808 & 20208 \\
            & Dev & 17447 & 912 & 1189\\
            & Test & 18284 & 580 & 872\\
    \midrule
        \multirow{2}{*}{MultiRC} 
            & Train & 23613 & 3630 & 0 \\
            & Dev & 4156 & 692 & 0\\
    \midrule
        \multirow{3}{*}{DREAM} 
            & Train & 16708 & 1530 & 110 \\
            & Dev & 5531 & 531 & 58\\
            & Test & 5588 & 495 & 40\\
    \midrule
        \multirow{2}{*}{CosmosQA} 
            & Train & 7298 & 848 & 32 \\
            & Dev & 60009 & 10889 & 400\\
    \bottomrule
    \end{tabular}
    }
    \caption{The proportion (absolute numbers) of neural, rule-based and Q+A examples in the hybrid datasets.}
    \label{tab:hybrid_splits}
\end{table}

%% file: appendix/conversion_examples.tex
Tables \ref{tab:example_conversions_race}, \ref{tab:example_conversions_multirc} and \ref{tab:example_conversions_dream} show examples of rule-based and neural conversions on RACE, MultiRC and DREAM respectively.

\begin{table*}[ht!]
\small
    \centering
    \begin{tabular}{p{0.3 \textwidth}|p{0.3\textwidth}|p{0.3 \textwidth}}
        & \textbf{Rule-based} & \textbf{Neural} \\ \bottomrule
        \textbf{Q:} How do suburban commuters travel to and from the city in Copenhagen at present?
            \break \textbf{A:} About one third of the suburban commuters travel by bike.
            & Suburban commuters travel to about one third of the suburban commuters travel by bike and from the city in Copenhagen at present.
            & Suburban commuters travel to and from the city in Copenhagen at present by bike \\
        \hline
        \textbf{Q:} What's the best title of the passage?
            \break \textbf{A:} Blame! Blame! Blame!
            & The best title of the passage's blame. 
            & The best title of the passage is Blame! Blame! blame! blamage!  \\
        \hline
        \textbf{Q:} What influence did the experiment have on Alexander ?
            \break \textbf{A:} He realized that slowing down his life speed could bring him more content.
            & The experiment had he realized that slowing down his life speed could bring him more content on Alexander.
            & The experiment influenced Alexander to realize that slowing down his life speed could bring him more content. \\
        \hline
       \textbf{Q:} Which of the following is TRUE about the report findings?
            \break \textbf{A:} The reading scores among older children have improved.
            & The reading scores among older children have improved is TRUE.
            & It is true that the reading scores among older children have improved. \\
        \bottomrule
    \end{tabular}
    \caption{Examples of Rule-based and Neural Conversions on RACE. }
    \label{tab:example_conversions_race}
\end{table*}

\begin{table*}[t!]
\small
    \centering
    \begin{tabular}{p{0.3 \textwidth}|p{0.3\textwidth}|p{0.3 \textwidth}}
        & \textbf{Rule-based} & \textbf{Neural} \\ \bottomrule
         \textbf{Q:} Timothy likes to spend his time after school doing what and with who?
            \break \textbf{A:} Timothy likes to play sports.
            & Timothy likes to spend his time after school doing what and with Timothy likes to play sports.
            & Timothy likes to play sports after school. \\
        \hline
        \textbf{Q:} What building were the four captives inside on Tuesday?
            \break \textbf{A:} CNN headquarters
            & The four captives inside on Tuesday were CNN headquarters.
            & The four captives were inside CNN headquarters on Tuesday. \\
        \hline
        \textbf{Q:} How might Air New Zealand's video partner benefited from helping to make this video?
            \break \textbf{A:} Coincides with the 50th anniversary of Sports Illustrated's Swimsuit franchise
            & Air New Zealand's video partner might benefited from helping to make this video by coincides with the 50th anniversary of Sports Illustrated's Swimsuit franchise.
            & Air New Zealand's video partner benefited from helping to make this video because it coincides with the 50th anniversary of Sports Illustrated's Swimsuit franchise. \\
        \hline
        \textbf{Q:} Did Alexander set out to secure his northern fronts and was he able to accomplish this goal?
            \break \textbf{A:} Yes and yes.
            & $\langle$ Unable to Convert $\rangle$
            & Alexander set out to secure his northern fronts and was he able to accomplish this goal. \\
        \bottomrule
        
    \end{tabular}
    \caption{Examples of Rule-based and Neural Conversions on MultiRC}
    \label{tab:example_conversions_multirc}
\end{table*}

\begin{table*}[t!]
\small
    \centering
    \begin{tabular}{p{0.3 \textwidth}|p{0.3\textwidth}|p{0.3 \textwidth}}
        & \textbf{Rule-based} & \textbf{Neural} \\ \bottomrule
         \textbf{Q:} What is one method of treatment the dentist does NOT mention?
            \break \textbf{A:} doing a root canal
            & Doing a root canal is one method of treatment the dentist NOT mentions.
            & One method of treatment the dentist does NOT mention is doing a root canal. \\
        \hline
        \textbf{Q:} How often does the woman see her parents?
            \break \textbf{A:} Once a week.
            & The woman sees her parents once a week.
            & The woman sees her parents once a week. \\
        \hline
        \textbf{Q:} What does the man think of the woman's idea at first?
            \break \textbf{A:} He strongly opposes it.
            & The man thinks he strongly opposes it of the woman's idea at first.
            & The man strongly opposes the woman's idea at first. \\
        \hline
        \textbf{Q:} What does the man think of the teacher?
            \break \textbf{A:} She's from Asia.
            & The man thinks she's from Asia of the teacher.
            & The man thinks the teacher is from Asia. \\
        \bottomrule
        
    \end{tabular}
    \caption{Examples of Rule-based and Neural Conversions on DREAM}
    \label{tab:example_conversions_dream}
\end{table*}

%% file: main.bbl
\begin{thebibliography}{28}
\expandafter\ifx\csname natexlab\endcsname\relax\def\natexlab#1{#1}\fi

\bibitem[{Bowman et~al.(2015)Bowman, Angeli, Potts, and Manning}]{Bowman2015}
Samuel~R. Bowman, Gabor Angeli, Christopher Potts, and Christopher~D. Manning.
  2015.
\newblock \href {https://doi.org/10.18653/v1/d15-1075} {{A large annotated
  corpus for learning natural language inference}}.
\newblock \emph{Conference Proceedings - EMNLP 2015: Conference on Empirical
  Methods in Natural Language Processing}, pages 632--642.

\bibitem[{Clark et~al.(2018)Clark, Cowhey, Etzioni, Khot, Sabharwal, Schoenick,
  and Tafjord}]{clark2018think}
Peter Clark, Isaac Cowhey, Oren Etzioni, Tushar Khot, Ashish Sabharwal, Carissa
  Schoenick, and Oyvind Tafjord. 2018.
\newblock Think you have solved question answering? try arc, the ai2 reasoning
  challenge.
\newblock \emph{arXiv preprint arXiv:1803.05457}.

\bibitem[{Cooper et~al.(1996)Cooper, Crouch, Van~Eijck, Fox, Van~Genabith,
  Jaspars, Kamp, Milward, Pinkal, Poesio et~al.}]{cooper1996usingsemantic}
Robin Cooper, Dick Crouch, Jan Van~Eijck, Chris Fox, Johan Van~Genabith, Jan
  Jaspars, Hans Kamp, David Milward, Manfred Pinkal, Massimo Poesio, et~al.
  1996.
\newblock Using the framework, fracas: a framework for computational semantics.
\newblock Technical report, Technical Report LRE 62-051 D-16, The FraCaS
  Consortium.

\bibitem[{Demszky et~al.(2018)Demszky, Guu, and Liang}]{Demszky2018}
Dorottya Demszky, Kelvin Guu, and Percy Liang. 2018.
\newblock Transforming question answering datasets into natural language
  inference datasets.
\newblock \emph{ArXiv}, abs/1809.02922.

\bibitem[{Devlin et~al.(2018)Devlin, Chang, Lee, and
  Toutanova}]{devlin2018bert}
Jacob Devlin, Ming-Wei Chang, Kenton Lee, and Kristina Toutanova. 2018.
\newblock Bert: Pre-training of deep bidirectional transformers for language
  understanding.
\newblock \emph{arXiv preprint arXiv:1810.04805}.

\bibitem[{Falke et~al.(2019)Falke, Ribeiro, Utama, Dagan, and
  Gurevych}]{falke2019ranking}
Tobias Falke, Leonardo~FR Ribeiro, Prasetya~Ajie Utama, Ido Dagan, and Iryna
  Gurevych. 2019.
\newblock Ranking generated summaries by correctness: An interesting but
  challenging application for natural language inference.
\newblock In \emph{Proceedings of the 57th Annual Meeting of the Association
  for Computational Linguistics}, pages 2214--2220.

\bibitem[{Filippova and Altun(2013)}]{gsc}
Katja Filippova and Yasemin Altun. 2013.
\newblock \href {https://www.aclweb.org/anthology/D13-1155} {Overcoming the
  lack of parallel data in sentence compression}.
\newblock In \emph{Proceedings of the 2013 Conference on Empirical Methods in
  Natural Language Processing}, pages 1481--1491, Seattle, Washington, USA.
  Association for Computational Linguistics.

\bibitem[{Gu et~al.(2016)Gu, Lu, Li, and Li}]{gu2016copy}
Jiatao Gu, Zhengdong Lu, Hang Li, and Victor~O.K. Li. 2016.
\newblock \href {https://doi.org/10.18653/v1/P16-1154} {Incorporating copying
  mechanism in sequence-to-sequence learning}.
\newblock In \emph{Proceedings of the 54th Annual Meeting of the Association
  for Computational Linguistics (Volume 1: Long Papers)}, pages 1631--1640,
  Berlin, Germany. Association for Computational Linguistics.

\bibitem[{Hermann et~al.(2015)Hermann, Kocisky, Grefenstette, Espeholt, Kay,
  Suleyman, and Blunsom}]{cnn}
Karl~Moritz Hermann, Tomas Kocisky, Edward Grefenstette, Lasse Espeholt, Will
  Kay, Mustafa Suleyman, and Phil Blunsom. 2015.
\newblock \href
  {http://papers.nips.cc/paper/5945-teaching-machines-to-read-and-comprehend.pdf}
  {Teaching machines to read and comprehend}.
\newblock In C.~Cortes, N.~D. Lawrence, D.~D. Lee, M.~Sugiyama, and R.~Garnett,
  editors, \emph{Advances in Neural Information Processing Systems 28}, pages
  1693--1701. Curran Associates, Inc.

\bibitem[{Huang et~al.(2019)Huang, Bras, Bhagavatula, and
  Choi}]{huang2019cosmos}
Lifu Huang, Ronan~Le Bras, Chandra Bhagavatula, and Yejin Choi. 2019.
\newblock Cosmos qa: Machine reading comprehension with contextual commonsense
  reasoning.
\newblock \emph{arXiv preprint arXiv:1909.00277}.

\bibitem[{Kapanipathi et~al.(2020)Kapanipathi, Thost, Patel, Whitehead,
  Abdelaziz, Balakrishnan, Chang, Fadnis, Gunasekara, Makni, Mattei,
  Talamadupula, and Fokoue}]{kapanipathi2019infusing}
Pavan Kapanipathi, Veronika Thost, Siva~Sankalp Patel, Spencer Whitehead,
  Ibrahim Abdelaziz, Avinash Balakrishnan, Maria Chang, Kshitij Fadnis, Chulaka
  Gunasekara, Bassem Makni, Nicholas Mattei, Kartik Talamadupula, and Achille
  Fokoue. 2020.
\newblock Infusing knowledge into the textual entailment task using graph
  convolutional networks.
\newblock \emph{Proceedings of the AAAI Conference on Artificial Intelligence}.

\bibitem[{Khashabi et~al.(2018)Khashabi, Chaturvedi, Roth, Upadhyay, and
  Roth}]{multirc}
Daniel Khashabi, Snigdha Chaturvedi, Michael Roth, Shyam Upadhyay, and Dan
  Roth. 2018.
\newblock \href {https://doi.org/10.18653/v1/N18-1023} {Looking beyond the
  surface: A challenge set for reading comprehension over multiple sentences}.
\newblock In \emph{Proceedings of the 2018 Conference of the North {A}merican
  Chapter of the Association for Computational Linguistics: Human Language
  Technologies, Volume 1 (Long Papers)}, pages 252--262, New Orleans,
  Louisiana. Association for Computational Linguistics.

\bibitem[{Khot et~al.(2018)Khot, Sabharwal, and Clark}]{khot2018scitail}
Tushar Khot, Ashish Sabharwal, and Peter Clark. 2018.
\newblock Scitail: A textual entailment dataset from science question
  answering.
\newblock In \emph{Thirty-Second AAAI Conference on Artificial Intelligence}.

\bibitem[{Kry{\'s}ci{\'n}ski et~al.(2019)Kry{\'s}ci{\'n}ski, McCann, Xiong, and
  Socher}]{kryscinski2019evaluating}
Wojciech Kry{\'s}ci{\'n}ski, Bryan McCann, Caiming Xiong, and Richard Socher.
  2019.
\newblock Evaluating the factual consistency of abstractive text summarization.
\newblock \emph{arXiv preprint arXiv:1910.12840}.

\bibitem[{Lai et~al.(2017{\natexlab{a}})Lai, Bisk, and
  Hockenmaier}]{lai-etal-2017-natural}
Alice Lai, Yonatan Bisk, and Julia Hockenmaier. 2017{\natexlab{a}}.
\newblock \href {https://www.aclweb.org/anthology/I17-1011} {Natural language
  inference from multiple premises}.
\newblock In \emph{Proceedings of the Eighth International Joint Conference on
  Natural Language Processing (Volume 1: Long Papers)}, pages 100--109, Taipei,
  Taiwan. Asian Federation of Natural Language Processing.

\bibitem[{Lai et~al.(2017{\natexlab{b}})Lai, Xie, Liu, Yang, and
  Hovy}]{lai2017race}
Guokun Lai, Qizhe Xie, Hanxiao Liu, Yiming Yang, and Eduard~H. Hovy.
  2017{\natexlab{b}}.
\newblock Race: Large-scale reading comprehension dataset from examinations.
\newblock In \emph{EMNLP}.

\bibitem[{Lewis et~al.(2019)Lewis, Liu, Goyal, Ghazvininejad, Mohamed, Levy,
  Stoyanov, and Zettlemoyer}]{bart}
Mike Lewis, Yinhan Liu, Naman Goyal, Marjan Ghazvininejad, Abdelrahman Mohamed,
  Omer Levy, Ves Stoyanov, and Luke Zettlemoyer. 2019.
\newblock \href {http://arxiv.org/abs/1910.13461} {Bart: Denoising
  sequence-to-sequence pre-training for natural language generation,
  translation, and comprehension}.

\bibitem[{Liu et~al.(2019)Liu, Ott, Goyal, Du, Joshi, Chen, Levy, Lewis,
  Zettlemoyer, and Stoyanov}]{liu2019roberta}
Yinhan Liu, Myle Ott, Naman Goyal, Jingfei Du, Mandar Joshi, Danqi Chen, Omer
  Levy, Mike Lewis, Luke Zettlemoyer, and Veselin Stoyanov. 2019.
\newblock Roberta: A robustly optimized bert pretraining approach.
\newblock \emph{arXiv preprint arXiv:1907.11692}.

\bibitem[{Qi et~al.(2018)Qi, Dozat, Zhang, and Manning}]{stanfordnlp}
Peng Qi, Timothy Dozat, Yuhao Zhang, and Christopher~D. Manning. 2018.
\newblock \href {https://nlp.stanford.edu/pubs/qi2018universal.pdf} {Universal
  dependency parsing from scratch}.
\newblock In \emph{Proceedings of the {CoNLL} 2018 Shared Task: Multilingual
  Parsing from Raw Text to Universal Dependencies}, pages 160--170, Brussels,
  Belgium. Association for Computational Linguistics.

\bibitem[{Rajpurkar et~al.(2016)Rajpurkar, Zhang, Lopyrev, and
  Liang}]{rajpurkar2016squad}
Pranav Rajpurkar, Jian Zhang, Konstantin Lopyrev, and Percy Liang. 2016.
\newblock Squad: 100, 000+ questions for machine comprehension of text.
\newblock In \emph{EMNLP}.

\bibitem[{Sun et~al.(2019)Sun, Yu, Chen, Yu, Choi, and Cardie}]{sun2019dream}
Kai Sun, Dian Yu, Jianshu Chen, Dong Yu, Yejin Choi, and Claire Cardie. 2019.
\newblock Dream: A challenge data set and models for dialogue-based reading
  comprehension.
\newblock \emph{Transactions of the Association for Computational Linguistics},
  7:217--231.

\bibitem[{Trivedi et~al.(2019)Trivedi, Kwon, Khot, Sabharwal, and
  Balasubramanian}]{Trivedi2019}
Harsh Trivedi, Heeyoung Kwon, Tushar Khot, Ashish Sabharwal, and Niranjan
  Balasubramanian. 2019.
\newblock Repurposing entailment for multi-hop question answering tasks.
\newblock In \emph{Proceedings of the 2019 Conference of the North American
  Chapter of the Association for Computational Linguistics: Human Language
  Technologies, Volume 1 (Long and Short Papers)}, pages 2948--2958.

\bibitem[{Wang et~al.(2020)Wang, Cho, and Lewis}]{wang2020asking}
Alex Wang, Kyunghyun Cho, and Mike Lewis. 2020.
\newblock Asking and answering questions to evaluate the factual consistency of
  summaries.
\newblock \emph{arXiv preprint arXiv:2004.04228}.

\bibitem[{Wang and Jiang(2015)}]{wang2015learning}
Shuohang Wang and Jing Jiang. 2015.
\newblock Learning natural language inference with lstm.
\newblock \emph{arXiv preprint arXiv:1512.08849}.

\bibitem[{Welleck et~al.(2019)Welleck, Weston, Szlam, and Cho}]{dnli}
Sean Welleck, Jason Weston, Arthur Szlam, and Kyunghyun Cho. 2019.
\newblock Dialogue natural language inference.
\newblock In \emph{ACL}.

\bibitem[{Williams et~al.(2018)Williams, Nangia, and Bowman}]{mnli}
Adina Williams, Nikita Nangia, and Samuel Bowman. 2018.
\newblock \href {http://aclweb.org/anthology/N18-1101} {A broad-coverage
  challenge corpus for sentence understanding through inference}.
\newblock In \emph{Proceedings of the 2018 Conference of the North American
  Chapter of the Association for Computational Linguistics: Human Language
  Technologies, Volume 1 (Long Papers)}, pages 1112--1122. Association for
  Computational Linguistics.

\bibitem[{Wolf et~al.(2019)Wolf, Debut, Sanh, Chaumond, Delangue, Moi, Cistac,
  Rault, Louf, Funtowicz, and Brew}]{huggingfaceTransformers}
Thomas Wolf, Lysandre Debut, Victor Sanh, Julien Chaumond, Clement Delangue,
  Anthony Moi, Pierric Cistac, Tim Rault, R'emi Louf, Morgan Funtowicz, and
  Jamie Brew. 2019.
\newblock Huggingface's transformers: State-of-the-art natural language
  processing.
\newblock \emph{ArXiv}, abs/1910.03771.

\bibitem[{Zhang et~al.(2020)Zhang, Zhang, and Manning}]{FactCCAnalysis}
Yuhui Zhang, Yuhao Zhang, and Christopher~D. Manning. 2020.
\newblock A close examination of factual correctness evaluation in abstractive
  summarization.

\end{thebibliography}
